%% file: main.tex
\documentclass[10pt,twocolumn,letterpaper]{article}
\usepackage{graphicx}
\usepackage{cvpr}              

\usepackage{graphicx}
\usepackage{amsmath}
\usepackage{amssymb}
\usepackage{booktabs}
\usepackage{times}
\usepackage{epsfig}
\usepackage{graphicx}
\usepackage{mmstyle}
\usepackage{bm}

\usepackage{algorithm}
\usepackage{algpseudocode}
\usepackage{amsmath}
\usepackage{graphics}
\usepackage{epsfig}

\usepackage[pagebackref,breaklinks,colorlinks]{hyperref}

\usepackage[capitalize]{cleveref}
\crefname{section}{Sec.}{Secs.}
\Crefname{section}{Section}{Sections}
\Crefname{table}{Table}{Tables}
\crefname{table}{Tab.}{Tabs.}

\def\cvprPaperID{*****} 
\def\confName{ICCV}
\def\confYear{2023}

\input{macros}

\begin{document}

\title{Augment Before Copy-Paste: Data and Memory Efficiency-Oriented \\
Instance Segmentation Framework for Sport-scenes}
\author{\textsuperscript{1 }Chih-Chung Hsu \quad \textsuperscript{2 }Chia-Ming Lee \quad \textsuperscript{3}Ming-Shyen Wu\\
	Institute of Data Science, National Cheng Kung University\\
	No.1, University Rd. Tainan City, Taiwan\\
	{\tt\small \textsuperscript{1}cchsu@gs.ncku.edu.tw \quad \textsuperscript{2}zuw408421476@gmail.com
 \quad \textsuperscript{3}
 Wu.Ming.Shyen@gmail.com}
}
\maketitle

\input{articles/abstract.tex}

\input{articles/introduction.tex}
\input{articles/methodology.tex}
\input{articles/experiment.tex}
\input{articles/conclusion.tex}

{\small
\bibliographystyle{ieee_fullname}
\bibliography{main}
}

\end{document}


\title{Supplementary Materials Template: }

\author{First Author\\
Institution1\\
Institution1 address\\
{\tt\small firstauthor@i1.org}
\and
Second Author\\
Institution2\\
First line of institution2 address\\
{\tt\small secondauthor@i2.org}
}

\maketitle

\section{Supplementary \#1}

\section{Supplementary \#2}

{\small
	\bibliographystyle{../ieee_fullname}
	\bibliography{../main}
}

%% file: macros.tex
\usepackage{xcolor}
\usepackage{wrapfig}
\usepackage{soul}

\definecolor{yellow}{rgb}{1,1, 0.6}
\definecolor{lightyellow}{rgb}{1,1, 0.8}
\definecolor{orange}{rgb}{1, 0.8, 0.6}
\definecolor{coral}{RGB}{246,131,65}
\definecolor{pinkred}{rgb}{1, 0.6, 0.6}
\definecolor{hotpink}{RGB}{238,64,195}
\definecolor{lavender}{RGB}{207,226,243}
\definecolor{gainsboro}{RGB}{208,224,227}
\definecolor{gainsboro2}{RGB}{217,234,211}
\definecolor{blanchedalmond}{RGB}{252,229,205}


%% file: articles/abstract.tex

\begin{abstract}
Instance segmentation is a fundamental task in computer vision with broad applications across various industries. In recent years, with the proliferation of deep learning and artificial intelligence applications, how to train effective models with limited data has become a pressing issue for both academia and industry. In the Visual Inductive Priors challenge (VIPriors2023), participants must train a model capable of precisely locating individuals on a basketball court, all while working with limited data and without the use of transfer learning or pre-trained models. We propose Memory effIciency inStance Segmentation framework based on visual inductive prior flow propagation that effectively incorporates inherent prior information from the dataset into both the data preprocessing and data augmentation stages, as well as the inference phase. Our team (ACVLAB)  experiments demonstrate that our model achieves promising performance (0.509 AP@0.50:0.95) even under limited data and memory constraints.
\end{abstract}

%% file: articles/introduction.tex

\section{Introduction}
\label{sec:introduction}

Instance segmentation, a cornerstone of computer vision within deep learning, boasts diverse applications like pedestrian detection and multi-object tracking. With the rise of deep learning, industries are integrating it into various aspects of their businesses to gain a competitive edge. However, implementing deep learning at these granular levels faces challenges such as insufficient annotated data and limited computational resources. These constraints can lead to models performing below expectations. Therefore, effectively leveraging limited data to train models has become one of today's research hotspots. The 'VIPriors: Visual Inductive Priors for Data-Efficient Deep Learning' workshop \cite{VIP2th}\cite{VIP3th} introduces a challenge, urging participants to build generalizable models by fusing dataset-specific prior knowledge in resource-scarce contexts. Notably, the use of pre-trained models is disallowed.

In the instance segmentation track, the goal is to predict players, basketballs, referees and coaches in a basketball court. Task-specific data augmentation based on instance traits yields substantial performance gains, validated in the challenge. Nevertheless, conventional state-of-the-art\cite{2021sota}\cite{2022sota} prioritize performance without due regard for computational resources. We propose a "augmentation before copy-paste" pipeline with RGB distortion and geometry transformation before copy-paste augmentation, exploiting object-derived semantic representation effectively.

Furthermore, we remove redundant image area to find out a region of interests. Utilizing prior knowledge from basketball court backgrounds—often defined by rectangular boundary lines, we extract court areas for training and inference on smaller images to minimize the computational consumption without feature distortion.




\begin{figure*}
    \centering
    \includegraphics[width=1\textwidth]{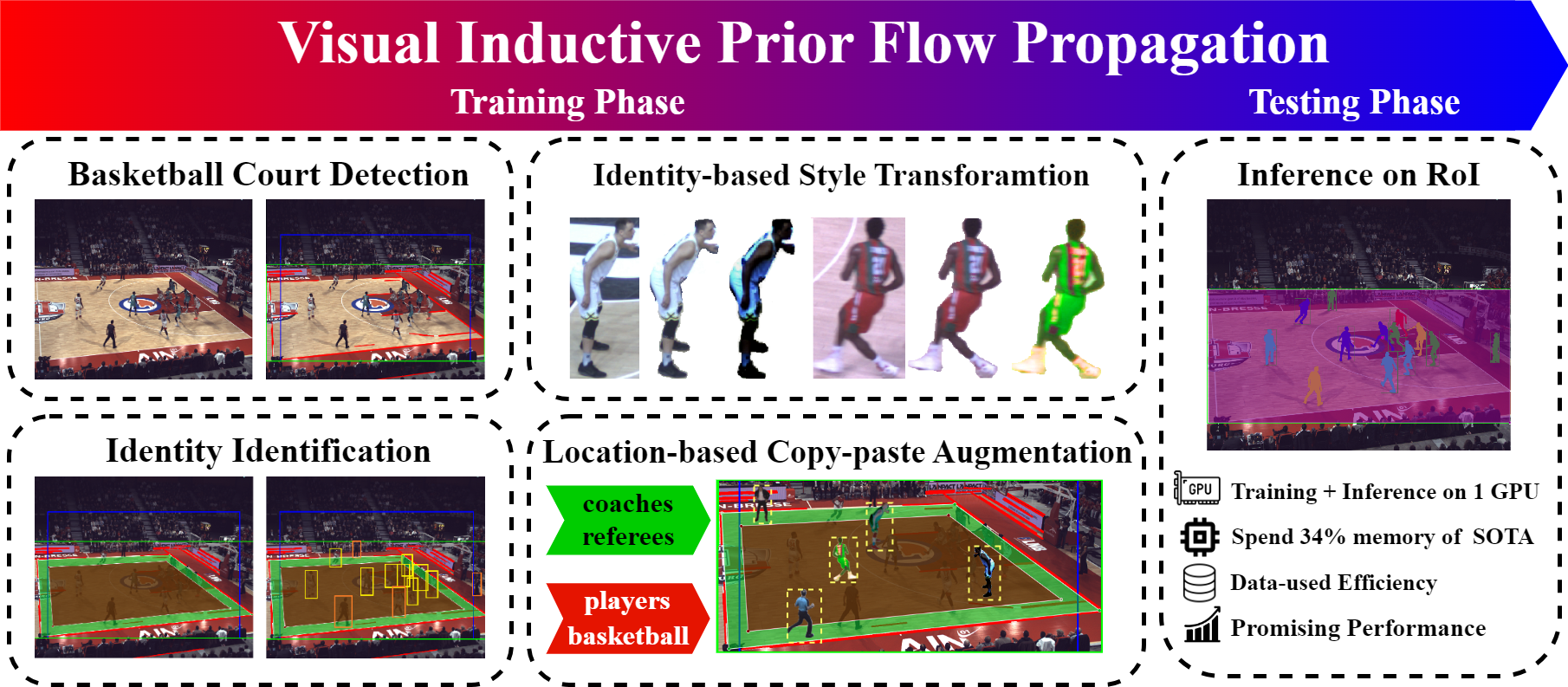}
    \caption{The overall  of proposed instance segmentation framework. The visual inductive prior is fully utilized at each stage to make effective optimizations. This approach not only reduces computational resource consumption but also maintains solid model performance. We begin by employing the Canny-Hough operator to adaptively combine image-level prior to detect the basketball court's position. Subsequently, we leverage class-level prior for identity identification. We then utilize this information for style transformation of various objects, integrating image-level prior knowledge through copy-paste augmentation. Finally, model inference solely based on the detected basketball court's location.}
    \label{fig:overall.png}	
\end{figure*}

In recent years, deep learning-based instance segmentation methods have garnered widespread recognition, exemplified by approaches such as Cascade-MaskRCNN\cite{CascadeRCNN}, Maskformer\cite{Maskformer}, queryinst\cite{QueryInst}, and CBNet\cite{CBNet}, among others. Over the past couple of years, \cite{2021sota}\cite{2022sota}. have demonstrated remarkable achievements within the VIPriors Workshop through the CBNet-based model architectures.

CBNet is distinguished by its capability to synergistically couple multiple backbone networks and detectors, enabling the effective fusion of both low-level and high-level semantic representations. This architecture, characterized by its scalability and ease of training, imparts a diverse range of model inductive attributes. It accomplishes this while maintaining noteworthy detection precision and generalization ability without compromising inference speed.

Data augmentation constitutes a critical facet in training deep learning models, especially when confronted with limited data. By subjecting original images to operations such as hue variations, geometric transformations, and erasure transformations, the diversity of features is augmented, enhancing the model's capacity for generalization across unseen domains.

Copy-paste augmentation\cite{CopyPaste} emerges as an effective augmentation strategy for instance segmentation task. It leverages prior knowledge of objects to enhance a model's generalization and robustness to out-of-domain objects. In VIPriors instance segmentation challenge , \cite{2021sota}\cite{2022sota} introduced task-specific copy-paste augmentation. This procedure leverages object-derived visual prior knowledge to optimize the data pipeline, ensuring that the features of pasted objects align more closely with plausible scenarios.
For instance, constraints can be applied to the coordinates of pasted images based on the probable player positions, effectively filtering out implausible results.


The experimental results confirm that our model can be effectively trained on a single GPU with 24 GB of memory while maintaining promising performance.

%% file: articles/methodology.tex

\section{Methodology}

In this section, we elaborate on all the components presented in Figure \ref{fig:overall.png}, including the basketball court detection algorithm, illustration about augmentation pipeline, and inference on region of interests.
\label{sec:methodology}


\subsection{Basketball Court Detection and Cropping}

Images with large sizes will prolong training and inference times, and may lead to memory 
insufficient. Resizing to a fixed size is common strategy to handle it. But it may lead to distorting features or losing image texture details. Conversely, the cropping approach can retain more information from source images. This method relies more heavily on prior knowledge within the image to determine the exact cropping boundaries. We introduce a basketball court detection and cropping algorithm, which is based on prior with canny-hough straight line detection operator\cite{Canny}\cite{Hough} to detect the location of basketball court and reduce image size.
\begin{algorithm}
\caption{Basketball Court Detection Algorithm}
\begin{algorithmic}[1]
\State \textbf{Data:} All image data denoted as $\textbf{I}_{\text{original}}$, All Cropped image data denoted as $\textbf{I}_{\text{cropped}}$
\State Denote $\phi(\cdot)$ as canny operator, and $\tau(\cdot)$ as hough operator
\For{each image $\textbf{I}_i$ in $\textbf{I}_{\text{original}}$}
    \State Initialize $\textbf{I}_{\text{ih}}$, $\textbf{I}_{\text{iw}}$ = the height and width of image $\textbf{I}_i$
    \State $\text{min}_{\text{h}} = \frac{1}{9} \textbf{I}_{\text{ih}}$, $\text{max}_{\text{h}} = \frac{8}{9} \textbf{I}_{\text{ih}}$, \State$\text{min}_{\text{w}} = \frac{1}{15} \textbf{I}_{\text{iw}}$, $\text{max}_{\text{w}} = \frac{14}{15} \textbf{I}_{\text{iw}}$
    \State Detect all lines $L$ in images $\tau(\phi(\textbf{I}_i))$
    \State Compute the maximum convex hull $\delta$ in $L$
    \State Crop $\textbf{I}_i$ based on the coordinate $(x, y, w, h) =$\\
    \hspace{\algorithmicindent}(min($\text{min}_{\text{w}}$, $\delta_x$), max($\text{min}_{\text{h}}$, $\delta_y$) - 50,\\
    \hspace{\algorithmicindent}max($\text{min}_{\text{w}}$, $\delta_w$), min($\text{max}_{\text{h}}$, $\delta_h$))
\EndFor
\end{algorithmic}
\end{algorithm}

\begin{figure}
    \centering
    \includegraphics[width=0.48\textwidth]{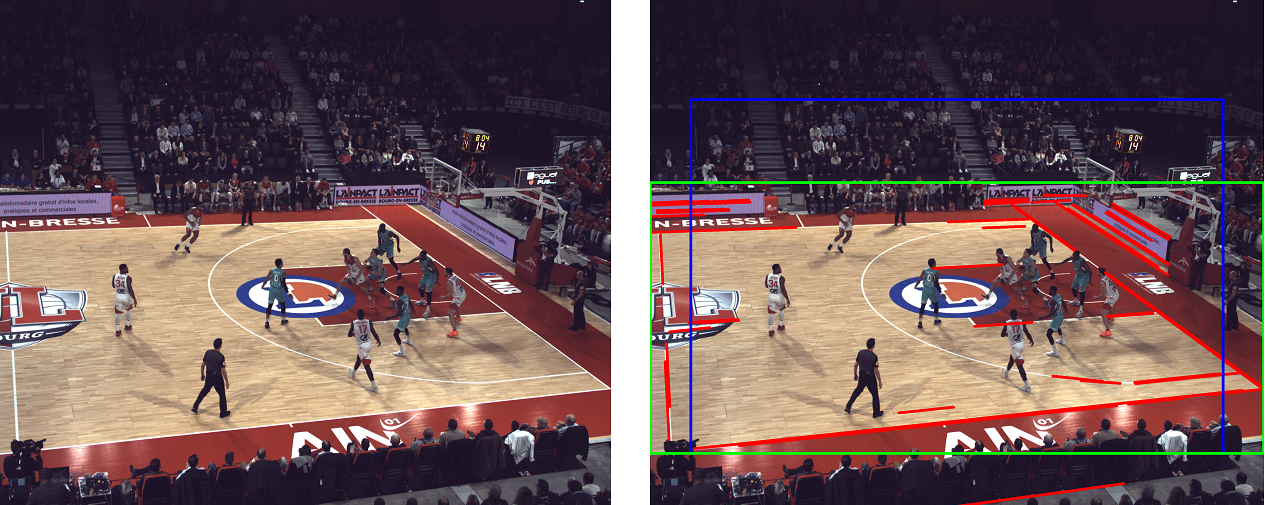}
    \caption{The illustrations for the cropping algorithm. The left figure is the original image. The right one is cropped, with red lines detected by the Canny edge detector and Hough transform. The blue line shows a boundary based on image size, while the green lines indicate dynamic boundary from the detected lines.}
    \label{fig:basketball detection and cropping.png}	
\end{figure}

\subsection{Identity Identification}

To further optimize the data augmentation pipeline, it is essential to make more effective use of prior knowledge. We have observed that referees and coaches generally stand around the perimeter of the basketball court for a better view of the players' movements, while the players themselves are active within the interior of the court. This information can be effectively utilized to estimate the likely identity of objects through basketball court area detection. Specifically, we consider 20\% of the detected area as a decision boundary and identify the objects based on the bottom coordinates of their detection boxes.

\begin{figure}
    \centering
    \includegraphics[width=0.48\textwidth]{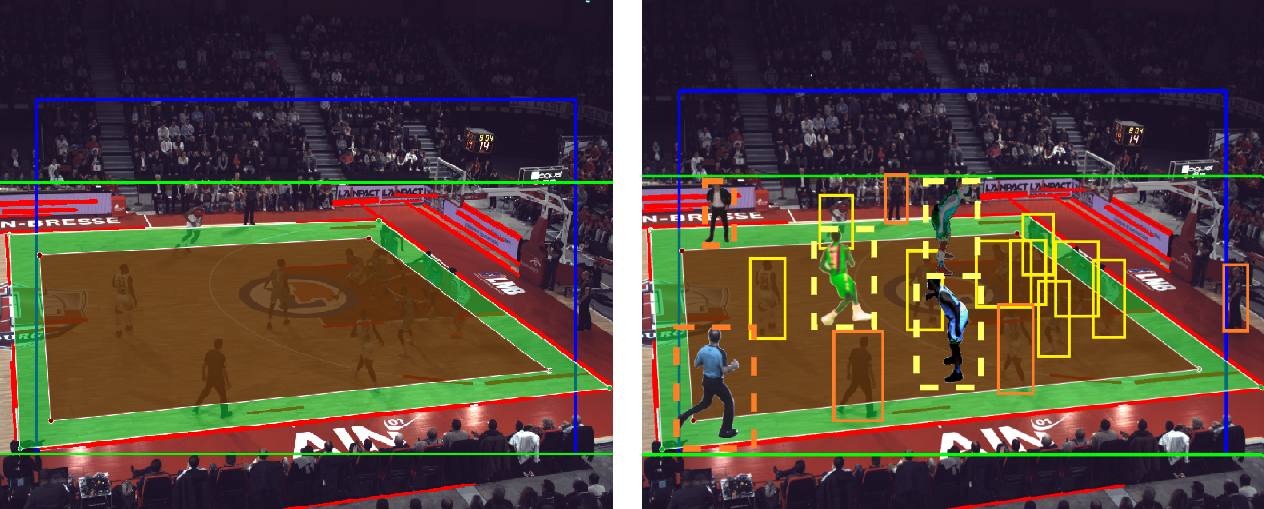}
    \caption{The left figure displays a region identified based on the maximum convex hull, which is determined using the endpoints of all lines detected by the Canny-Hough operator. The subclass attributes of the object are determined by its bounding box coordinates. The object marked by a dotted line represents the result of location-based copy-paste augmentation. }
    \label{fig:identity identification.png}
\end{figure}

\subsection{Identity-based Style Transformation}

Previous research in data augmentation presents two issues:(1) whole image-level augmentation may unnecessarily increase the complexity of the feature space, resulting in limited performance gains;(2) the data augmentation procedures for different classes or scenarios are incomplete. Objects on the basketball court are diverse. The 'human' class may have different sub-class, including 'player, referee, coach,' each with highly diverse internal feature attributes; the 'ball' class may also exist in various lighting and occlusion conditions, as well as variations in the game ball itself. Using a few classes to simply distinguish object attributes can make it difficult for basic data augmentation procedures to effectively expand the source domain. Further leveraging prior knowledge to incrementally decompose high-level classes can improve the model's predictive capabilities for unseen targets to a certain extent.

Specifically, we can distinguish the sub-classes of objects through the content explained in Section 3-2 and then apply different enhancement strategies to them. Players may wear clothing of different colors, high saturation, and strong contrast to increase their distinctiveness, or they may have varying skin tones or genders. For the 'player' sub-class, we employ RGB curve distortion for object-level data augmentation. As for other categories such as referees, coaches, and balls, where the available prior knowledge is relatively limited, we resort to using salt-and-pepper noise and brightness variations to increase the model's robustness against varying lighting conditions.

\begin{figure}
    \centering
    \includegraphics[width=0.5\textwidth]{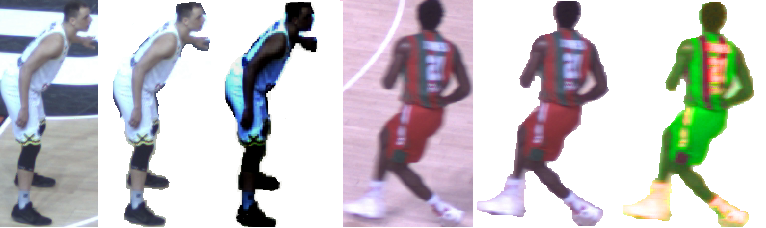}
    \caption{The demo of identity-based style transfer applied to basketball players. Significant variations in appearance are evident after the hue or RGB transformation. In the left example, there is a noticeable change in skin tone, while in the right example, the player's jersey changes dramatically, almost as if he has switched to a different team.}
    \label{fig:identity transform.png}
\end{figure}

\subsection{Location-based Copy-paste Augmentation}

In previous research on copy-paste augmentation, the coordinates of the objects are constrained within a range determined by the image's height and width, as denoted by the blue bounding box in Figure
\ref{fig:basketball detection and cropping.png}. Such restrictive boundaries might result in objects being augmented in unreasonable locations. In our study, we vary the possible boundary regions based on prior knowledge. As described in Sec. 3-2 and 3-3, our proposed algorithm can effectively determine the areas where various types of objects are likely to appear. We then perform copy-paste augmentation based on these locations.

\subsection{Inference on Region of Interests}

To better utilize memory usage and reduce computational consumption, we believe that just inferencing specific region via the vision prior to crop out redundant area like Sec.3-2, 3-4, is an efficient and effective way to achieve this goal.
We reduce memory usage and inference time during model inference by resizing images while preserving the region of interest.



\subsection{Model Architecture}

Our model is architecturally founded on the HybridTaskCascade  \cite{HTC} detector. Subsequent to this foundation, we utilize the CB-SwinTransformer-Base \cite{CBNet}\cite{swin} as our backbone to extract image features. After this, we integrate the CB-feature pyramid network, which employs group normalization \cite{groupnormalization} as model's neck to better capture  from low to high-level feature representations. As for model's head parts, we use the region proposal network with its default setting. This is further followed by the inclusion of the HybridTaskCascadeRoIHead. Within this RoI head, there are two pivotal components: the bounding-bbox head, which retains its default setting, and the mask-head. The mask-head is replaced by mask-scoring head \cite{MaskScoringRCNN} to improve model performance on instance's texture and boundary details.

%% file: articles/experiment.tex

\section{Experiments}
\label{sec:experiment}

\subsection{Training Details}

All our experiments are conducted on a single GPU (NVIDIA TITAN RTX) and are based on the MMDetection toolbox\cite{mmdetection}.

Our dataset is provided by Synergy Sports. On VIPriors instance segmentation challenge, there are 184, 62, 64 images in the training, validation, testing set. 

We train our model with totally 36 epoches. Within this framework, we adopt the AdamW optimizer, setting the learning rate at 0.0001 and the weight decay at 0.05. The batch size is set to 1 due to the limitation of GPU memory. We duplicate training and validation images 10 times, then implement proposed augmentation to train instance segmentation model.

\begin{figure}
    \centering
    \includegraphics[width=0.5\textwidth]{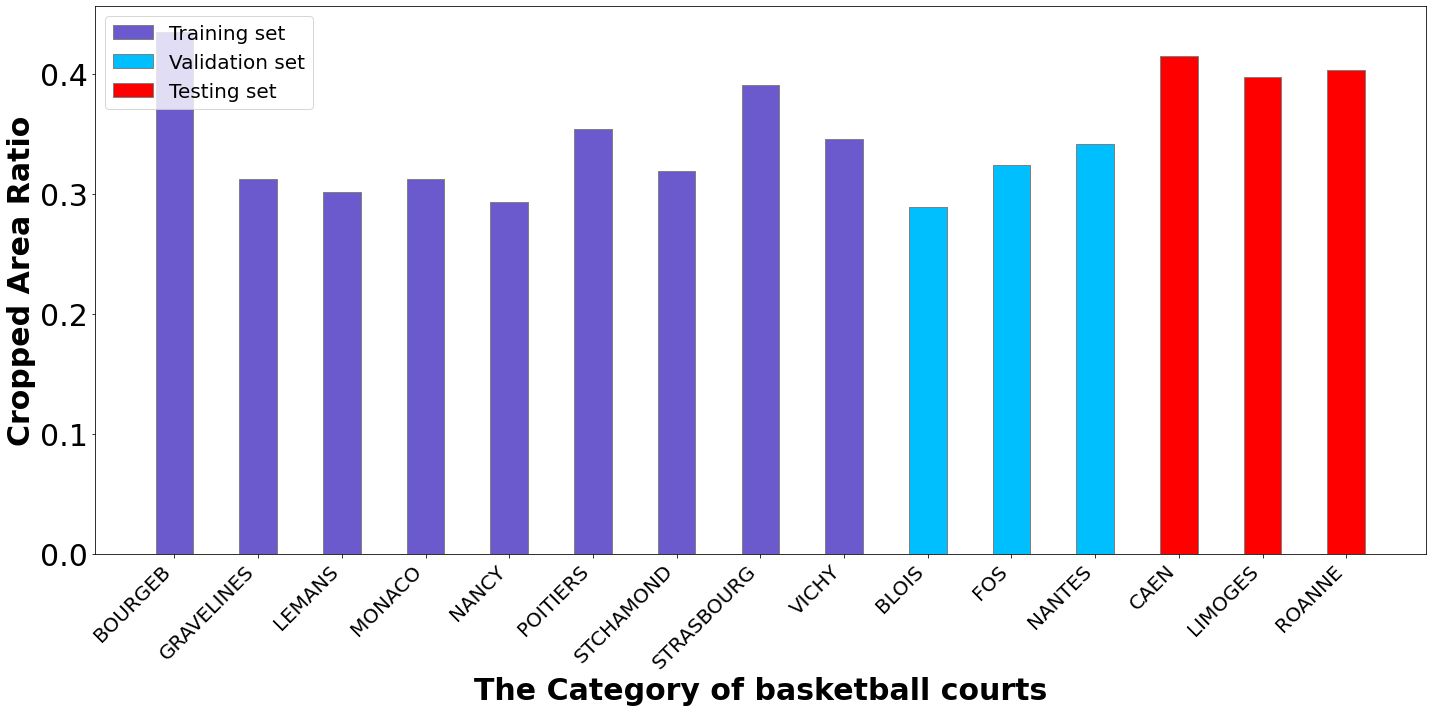}
    \caption{The cropped area statistic barchart. The x-axis is corresponding to basketball courts; the y-axis is the cropped area ratio against whole raw image. From left to right, the three colors correspond to the training, validation, and testing set.}
    \label{fig:cropoutcome.png}	
\end{figure}

After undergoing the cropping pipeline, the image sizes of the training set, validation set, and test set are reduced by 33.98 \%, 33.17 \%, and 40.72 \% of the original sizes, respectively. We conduct statistics for each basketball court category, as shown in Figure \ref{fig:cropoutcome.png}.

On the other hand, in the online augmentation, each training images have a 0.5 probability of undergoing a horizontal flip, and is then randomly resized to either (1400, 800) or (1400, 1200). Subsequently, 70\% of the image area is randomly cropped from it, and normalization operator is used on each images. Finally, GridMask\cite{Gridmask} augmentation is performed on each images.

\begin{table*}[]
\label{tab:ablationstudy}
\begin{tabular}{@{}lclclclclcl@{c}}
\toprule
\multicolumn{1}{c}{Models} & \multicolumn{1}{c}{AP@0.50}                        & \multicolumn{1}{c}{AP@0.50:0.95} & \multicolumn{1}{c}{\begin{tabular}[c]{@{}c@{}}AP@0.50:0.95\\ (small)\end{tabular}} & \multicolumn{1}{c}{\begin{tabular}[c]{@{}c@{}}AP@0.50:0.95\\ (medium)\end{tabular}} & \multicolumn{1}{c}{\begin{tabular}[c]{@{}c@{}}AP@0.50:0.95\\ (large)\end{tabular}} \\ \midrule
Vanillia instance segmentation model
&\multicolumn{1}{c}{0.789}&\multicolumn{1}{c}{0.403}&0.401&\multicolumn{1}{c}{0.470}&\multicolumn{1}{c}{0.631}\\ 
With simple copy-paste augmentation
&\multicolumn{1}{c}{0.863}&\multicolumn{1}{c}{0.444}&\multicolumn{1}{c}{0.462}&\multicolumn{1}{c}{0.561}&\multicolumn{1}{c}{0.667}\\ 
With proposed augmentation pipeline
&\multicolumn{1}{c}{0.870}&\multicolumn{1}{c}{\textbf{0.481}}&\multicolumn{1}{c}{0.515}&\multicolumn{1}{c}{0.579}&\multicolumn{1}{c}{0.700}\\ 
With post-processing &\multicolumn{1}{c}{\textbf{0.896}}&\multicolumn{1}{c}{0.509}&\multicolumn{1}{c}{0.533}&\multicolumn{1}{c}{0.584}&\multicolumn{1}{c}{0.731}\\ \bottomrule
\end{tabular}

\caption{Results of Ablation Study for VIPriors instance segmentation challenge 2023 with or without the proposed augmentation pipeline.}
\end{table*}
\subsection{Post-Processing}

During the training process, we faced constraints related to limited GPU memory. Consequently, we abstained from resizing the images to larger dimensions in both training and testing phases to extract more granular feature information. This compromise adversely impacted the performance of our model. To mitigate this drawback, we employed the Stochastic Weight Averaging\cite{zhang2020swa} strategy to obtain the average model weights over subsequent training epochs. Further, we applied variable-intensity GridMask and executed additional data augmentation techniques on the original dataset to generate diverse training samples. Subsequently, we leveraged model ensemble and modelsoup\cite{modelsoup} to enhance the overall performance of our model.

\subsection{Ablation Study and Performance Comparison}

Our experimental results are presented in Table 1 and 2. The outcomes demonstrate that our proposed method, based on vision inductive prior, can effectively surpass the performance established by conventional approaches. 

The model shows strong performance on the AP@0.50 metric, implying that it can effectively detect the majority of instances in the testing set. However, it underperforms in terms of fine-grained segmentation. As a result, the overall performance at the AP@0.50:0.95 metric is slightly below the state-of-the-art model \cite{2022sota}.
Finally, we achieved a final result of 0.509 on the AP@0.50:0.95 metric.


Additionally, our model not only requires significantly less memory usage compared to \cite{2022sota} , using only 34.6\% of its memory, but also maintains competitive performance.

\begin{table}[]
\resizebox{\columnwidth}{!}{%
\begin{tabular}{lcccc}
\hline
\multicolumn{1}{c}{Methods} &
  \begin{tabular}[c]{@{}c@{}}AP@\\ 0.50:0.95\end{tabular} &
  \begin{tabular}[c]{@{}c@{}}AP@\\ 0.50\end{tabular} &
  \begin{tabular}[c]{@{}c@{}}Memory \\ (G)\end{tabular} &
  \begin{tabular}[c]{@{}c@{}}Inference times\\ (s)\end{tabular}\\ \hline
Yunusov et al. \cite{2021sota}& 0.477          & 0.747          & 27.1           & 6.47 \\
Yan et al.     \cite{2022sota}& \textbf{0.531} & 0.837     & 65.6            & 6.98\\
\hline
\textbf{Ours}  & 0.509          & \textbf{0.896} & \textbf{22.7}  & \textbf{3.95}

\end{tabular}%
}
\caption{Comparison of performance and computational resource requirements. We compare proposed method with the SOTA from VIP2021, 2022. *The architectures of these methods are CBNet-based, but the training process or hyperparameters may vary.}
\label{tab:comparison}
\end{table}

%% file: articles/conclusion.tex

\section{Conclusion}
\label{sec:conclusion}

In this paper, we propose an efficient instance segmentation framework that integrates visual inductive priors into various stages, including data preprocessing, data augmentation, and model inference stage. The experiments demonstrate that such an approach can significantly enhance model performance even in resource-constrained environments, without the need for any pre-trained weight or the use of transfer learning. 

Notably, increasing the image size during the training and inference time can toward improve model performance if there is sufficient available memory.